\documentclass[11pt]{article}

\usepackage[final]{acl}

\usepackage{times}
\usepackage{latexsym}

\usepackage[T1]{fontenc}

\usepackage[utf8]{inputenc}

\usepackage{microtype}

\usepackage{inconsolata}

\usepackage{graphicx}
\usepackage{amsmath}
\usepackage{paralist}
\usepackage{booktabs}
\usepackage{caption}

\title{The Tatoxa System for Text Detoxification in Low-Resource Languages: The Case of Tatar}

\author{
 \textbf{Ilseyar Alimova\textsuperscript{1}},
 \textbf{Bogdan Monogov\textsuperscript{1}},
 \textbf{Artyom Mazur\textsuperscript{2}},
 \textbf{Daniil Antonov\textsuperscript{3}},
\\
 \textbf{Vsevolod Karimov\textsuperscript{1,2}},
 \textbf{Vitaliy Egorov\textsuperscript{1}},
 \textbf{Bulat Khakimov\textsuperscript{4,5}},
 \textbf{and Alexander Panchenko\textsuperscript{1,6}}
\\
\\
 \textsuperscript{1}Skoltech,
 \textsuperscript{2}HSE,
 \textsuperscript{3}ITMO,
 \textsuperscript{4}Institute of Applied Semiotics
Tatarstan Academy of Sciences,
\\
\textsuperscript{5}Kazan Federal University,
 \textsuperscript{6}AIRI \\
 \small{
   \textbf{Correspondence:} \href{mailto:alimovailseyar@gmail.com}{alimovailseyar@gmail.com}
 }
}

\begin{document}
\maketitle
\begin{abstract}

Text detoxification, the automated detection and mitigation of abusive and harmful content, is essential for ensuring the safety of online communities and protecting users. However, low resource languages such as Tatar have received little research attention. In this paper we present Tatoxa, a novel state-of-the-art system for text detoxification in the Tatar language. Comparative experiments show that the proposed approach outperforms existing open source and proprietary commercial LLMs on key quality metrics. We also introduce a new dataset for text detoxification in Tatar, designed for fine tuning and evaluation in low resource settings. Finally, cross lingual transfer experiments indicate that transfer from other languages, including the culturally close Russian, performs significantly worse than training on native Tatar data even when a large Russian corpus is available. 

\end{abstract}

\section{Introduction}

Text detoxification is the process of rewriting text into a more neutral form that does not contain insults, profanity, or aggression. Such a technique is valuable for moderating content on social media: instead of censoring via deletion of posts containing toxic or obscene language, automated rewriting can produce sanitized versions that preserve the original meaning, thereby enhancing the safety of online interaction. 

The Multilingual Text Detoxification Shared Task conducted in CLEF 2025 shows that the quality of automatic methods for text detoxification task is still far from human baseline, especially for low-resource languages~\cite{dementieva2025overview}. This limitation arises from the widespread use of a single multilingual large language model (LLM) across languages. Such models often underperform on low-resource languages and are therefore less effective at reliably rewriting texts in those languages. Moreover, text detoxification depends on cultural knowledge: without familiarity with the source community’s norms and pragmatic cues, even a human may fail to identify toxic content, and automated systems are correspondingly less reliable.

\begin{figure}[!t]
  \centering
  \includegraphics[width=\columnwidth]{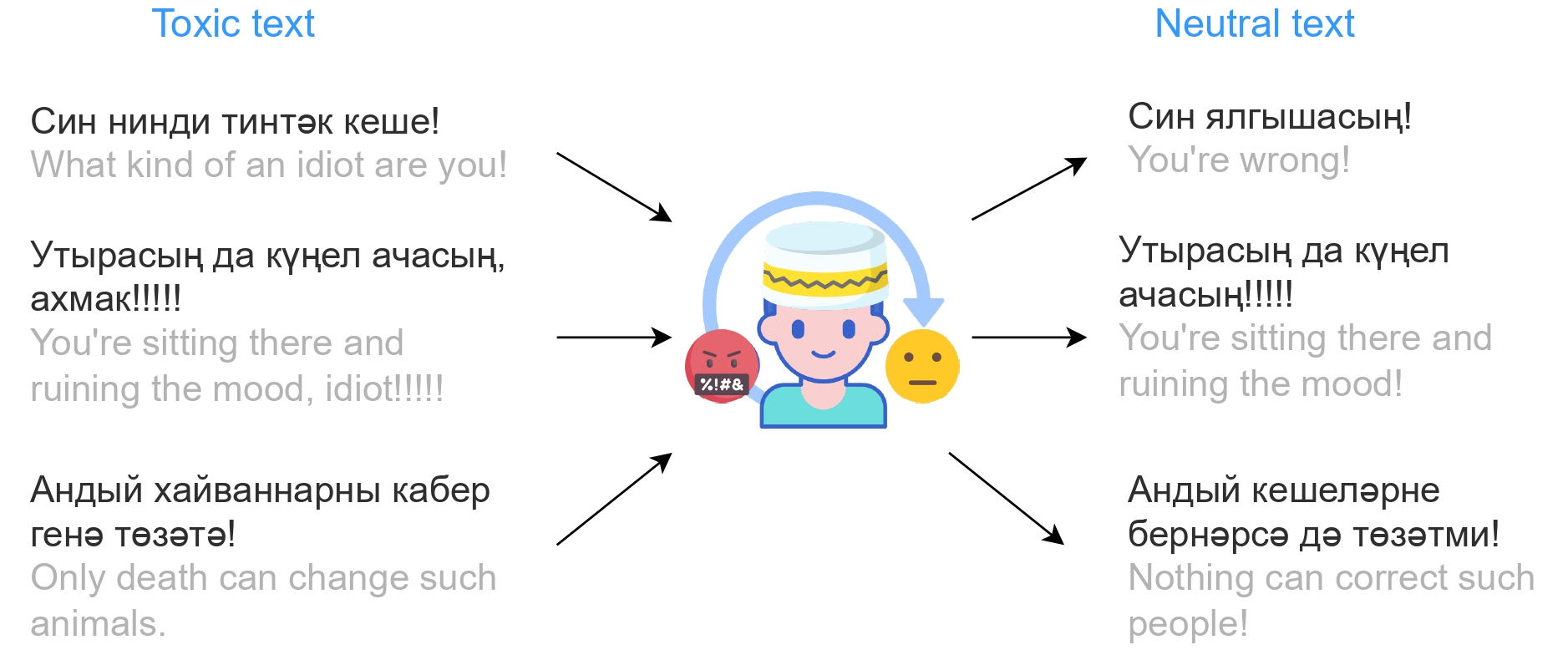}
  \caption{Example of the Tatoxa text detoxification for the Tatar language: original (left) and detoxified (right).}
  \label{fig:tatoxa}
\end{figure}


In this paper we investigate text-detoxification approaches for low-resource languages, using Tatar as a case study. CLEF-2025 TextDetox shared-task results indicate that automatic systems for Tatar achieved the lowest scores among the evaluated languages~\cite{dementieva2025overview}. To address this gap, we present Tatoxa, a new system adapted to Tatar, and evaluate state‑of‑the‑art methods for Tatar, demonstrating their effectiveness on the Tatar subset of the CLEF‑2025 detoxification dataset (CLEF‑Tatar)~\cite{dementieva2025overview}. An example of Tatoxa's output is shown in the Figure~\ref{fig:tatoxa}. We also manually augment the corpus with 701 manually annotated examples to support experiments with Tatar-only training and to study cross-lingual transfer, and we report improvements in task-specific metrics. The main contributions of this paper are:
\begin{itemize}
    \item We introduce Tatoxa, a new state-of-the-art method for text detoxification in the Tatar language. 
    \item We extend existing datasets by adding new toxic–non‑toxic text pairs in Tatar. 
    \item We conduct experiments on cross‑lingual data transfer to evaluate the portability of text detoxification methods across languages.
\end{itemize}

The remainder of this paper is organized as follows: Section~\ref{sec:related_work} reviews related work on detoxification and cross‑lingual transfer; Section~\ref{sec:tatoxa} describes Tatoxa; Section~\ref{sec:dataset} presents the dataset; Section~\ref{sec:experiments} outlines the experiments; Section~\ref{sec:results} reports results; Section~\ref{sec:conclusion} concludes; and Section~\ref{sec:limitations} discusses limitations.

\section{Related Work}
\label{sec:related_work}

While substantial progress in text detoxification has been achieved for high-resource languages such as English and Russian through the availability of parallel corpora ~\cite{dementievaMultiParaDetox2024}, multilingual coverage remains uneven, with many languages still lacking detoxification resources. MultiParaDetox extends the ParaDetox pipeline beyond English by introducing a scalable crowdsourcing framework that enables the collection of parallel detoxification data for additional 2 languages, illustrating that multilingual detoxification research spans both genuinely low-resource and moderately resourced settings ~\cite{dementievaUKR}.

Recent studies have begun to address this gap by proposing data-centric approaches tailored to specific linguistic and cultural contexts. For several African languages, a lightweight and interpretable pipeline combining TF–IDF–based toxicity detection with rule-based rewriting was introduced~\cite{detox_african}. The experiments demonstrated that hybrid, linguistically informed methods remain effective under extreme data scarcity. For Hebrew, the HeDetox corpus was constructed using few-shot large language model prompting followed by systematic human correction, showing that high-quality parallel detoxification data requires human verification even when LLMs are used for annotation~\cite{detox_hebrew}. Similarly, for Bengali, the large-scale BANGLANIRTOX corpus was developed via an LLM-assisted annotation pipeline, demonstrating that fine-tuned generative models outperform zero-shot prompting and translation-based baselines in end-to-end detoxification~\cite{detox_bengali}. Comparable findings have been reported for other underrepresented languages. For Italian, Detoxify-IT introduces the first parallel detoxification corpus and shows that even limited language-specific fine-tuning leads to clear improvements over zero-shot LLM prompting and generic multilingual baselines, reinforcing the importance of in-domain supervision for effective detoxification~\cite{detoxify_it}.

Recent work has also explored synthetic parallel data generation as a scalable alternative to manual annotation. SynthDetoxM demonstrates that modern open-source LLMs can act as effective few-shot annotators for creating multilingual parallel detoxification corpora, and shows that models fine-tuned on such synthetic data outperform zero-shot prompting and comparably sized human-annotated datasets, further reinforcing the central role of parallel supervision in data-scarce settings~\cite{synthdetoxm}.

Beyond text rewriting, related work has also explored toxicity detection and cross-lingual transfer for low-resource languages. For Ukrainian,~\cite{dementievaUKR} created the first toxicity classification corpus and evaluated methods such as back translation, adapter training, and LLM prompting, finding that fine-tuning on human-annotated, language-specific data yields the best performance, while cross-lingual transfer alone provides weaker baselines. Similar conclusions are drawn for several Indic languages, where manually verified multilingual datasets are shown to be crucial for building reliable safety models~\cite{detox_hindi}. Recent benchmark-oriented work further demonstrates that detoxification quality varies substantially across languages and evaluation metrics, with model rankings changing depending on the language and scoring setup, highlighting the limitations of toxicity-only automatic evaluation and motivating more linguistically grounded, multilingual evaluation protocols~\cite{nine_language_benchmark}.

Finally, despite these advances, text detoxification for Turkic languages, particularly Tatar, remains challenging. The CLEF-2025 Multilingual Text Detoxification shared task introduced the first fully human-annotated parallel detoxification dataset for Tatar, enabling systematic evaluation in this low-resource setting ~\cite{dementieva2025overview}. Results from the competition indicate that while fine-tuned and hybrid multilingual systems achieve the strongest overall performance across languages, Tatar stands out as one of the most difficult cases: notably, the overall winning system of the shared task did not achieve competitive results on Tatar, and the best performance relied on explicit, language-specific vocabulary adaptation rather than purely model-driven generation.

Overall, prior work consistently indicates that high-quality parallel supervision (human-annotated or carefully generated with LLM assistance) is the primary driver of successful detoxification in low-resource languages. Fine-tuning on in-domain data reliably outperforms zero-shot prompting and cross-lingual transfer alone, while hybrid and rule-guided approaches remain competitive in extremely resource-constrained scenarios and for languages with strong orthographic or cultural constraints.

\section{Tatoxa}
\label{sec:tatoxa}

\begin{figure*}[!t]
  \includegraphics[width=\textwidth]{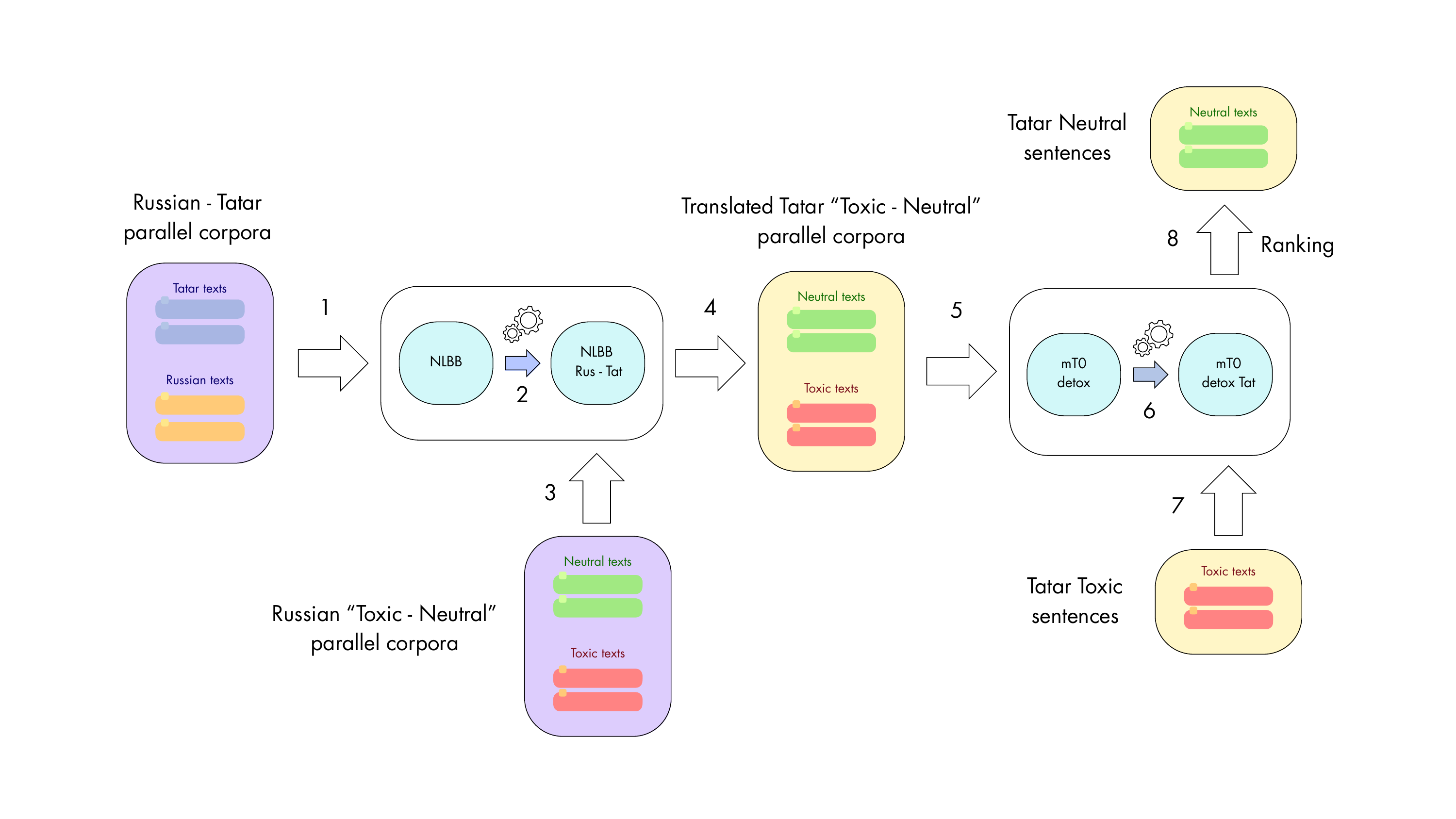}
  \caption{The diagram shows the Tatoxa pipeline workflow: (i) fine‑tuning a machine‑translation model to translate Russian into Tatar (steps 1–2); (ii) translating the detoxification dataset from Russian into Tatar (steps 3–4); (iii) fine‑tuning a detoxification model on Tatar data (steps 5–6); (iv) applying the detoxification model to Tatar texts and ranking candidate outputs to select the optimal result (steps 7–8).}
  \label{fig:tatoxa_pipeline}
\end{figure*}

Tatoxa leverages a large corpus translated from Russian into Tatar and follows a four-stage pipeline: 
\begin{inparaenum}[(i)]
\item fine-tuning a neural machine-translation model (NMT) for Russian $\rightarrow$ Tatar;
\item using the fine-tuned NMT to translate the detoxification dataset into Tatar;
\item  training a detoxification model on the translated dataset; 
\item performing inference with multi-candidate generation followed by ranking.
\end{inparaenum}
The resulting model is then applied to perform detoxification (toxicity mitigation) of texts in Tatar. The overall pipeline is presented in Figure~\ref{fig:tatoxa_pipeline}. The following provides a detailed description of each step. Source code and dataset are openly available.\footnote{\url{https://github.com/s-nlp/tatoxa}}

\subsection{Machine Translation Model}

To obtain higher-quality synthetic data, we first adapt a multilingual MT model to the Russian--Tatar language pair. We start from NLLB-200 model~\cite{nllbteam2022languageleftbehindscaling} and fine-tune it on the parallel corpus \href{https://huggingface.co/datasets/AigizK/tatar-russian-parallel-corpora}{Tatar-Russian parallel corpora}. The dataset contains aligned sentence pairs in Tatar and Russian. For each aligned pair we create two supervised training instances (Tatar→Russian and Russian→Tatar) so a single model is trained to translate in both directions. 

\subsection{Translating the Detoxification Dataset}

Since there is lack of parallel detoxification data in Tatar, we build a synthetic corpus by translating public Russian detoxification datasets. For each Russian pair of sentences we translate both sentences with the adapted on the previous stage NLLB-200 model. This yields synthetic Tatar detoxification pairs. For training split construction we use: 
\begin{inparaenum}[(i)]
    \item \href{https://huggingface.co/datasets/s-nlp/ru_paradetox}{Russian ParaDetox}~\cite{dementievaMultiParaDetox2024} corpus,
    \item Russian part of Multilingual  \href{https://huggingface.co/datasets/textdetox/multilingual_paradetox}{ParaDetox corpus}~\cite{dementieva2025overview},
    \item \href{https://huggingface.co/datasets/d0rj/rudetoxifier_data}{RuDetoxifier}
    \item \href{https://huggingface.co/datasets/d0rj/toxic_dvach_detoxified}{Detoxified} corpus.
\end{inparaenum}
Translation-based synthesis introduces noise due to imperfections in machine translation. To mitigate this, we filter synthetic sentence pairs using cross-lingual semantic similarity. We embed the Russian originals and their Tatar translations into a shared vector space with LaBSE~\cite{feng2022language}. For each example, we compute the cosine similarity between the Russian sentence and its Tatar translation separately for the toxic and neutral utterances, and retain the example only if both similarity scores are at least 0.7. Presented threshold chosen based on the empirical studies published in \cite{threshold}.  After filtering, the dataset contains 38,380 parallel pairs, of which 31,218 are used for training and 7,162 for validation.

\subsection{Text Detoxification Model}

For the final detoxification model, we used the mT0-XL model~\cite{muennighoff2023crosslingual} trained on automatically translated pairs from the previous step.To improve robustness we train an ensemble of LoRA adapters using $K$-fold splitting with $K{=}3$. We split the filtered training set into 3 folds, for each fold $k \in \{1,2,3\}$ we train an adapter $A_k$ on the corresponding training partition and evaluate on the held-out fold to select the best checkpoint per fold based on validation loss. All adapters share the same frozen mT0-XL backbone, only LoRA weights differ across folds.


\subsection{Inference and Candidate Ranking}

Single-shot generation can either insufficiently detoxify (leaving residual toxicity) or excessively detoxify (resulting in semantic drift). Therefore, we applied a strategy that generates multiple detoxified candidate sentences and then ranks the resulting candidates according to two criteria: their level of neutrality and their semantic similarity to the original sentence. For neutrality scoring we used an XLM-R based toxicity classifier~\cite{dementieva2025overview} semantic similarity between the original and detoxified text was measured with LaBSE~\cite{feng2022language}. For each adapter, we generate 60 candidates, yielding 180 candidates in total. We then rank all candidates by neutrality and semantic similarity and select the one with the highest combined score.

\section{Dataset}
\label{sec:dataset}

Given the relatively small size of the Tatar part for CLEF-2025 competition dataset, we expanded it by adding 701 examples. These were curated to be as consistent as possible with the original dataset and its annotation scheme. The source toxic examples was obtained from Tatar part of Multilingual Toxicity Dataset~\cite{dementievaMultiParaDetox2024}. This dataset provides examples for binary classification task to find whether the example is toxic or not. he dataset’s source material was drawn from a corpus of user-generated content on social media platforms.

\subsection{Annotation Methodology}

The annotation was carried out by two annotators and subsequently reviewed by a moderator. The annotators followed the guidelines provided by the CLEF‑2025 organizers. The main task was to minimally detoxify the text with making as few changes as possible while preserving the original meaning. Both annotators and the moderator are native speakers, the moderator additionally holds qualifications in natural language processing (NLP). In addition to the main annotations, annotators were asked to indicate the type of changes made to the text during detoxification: deletion or rewriting. Deletion denotes removing the toxic word while leaving the remainder of the text unchanged. Rewriting indicates that part of, or possibly the entire, original sentence was reformulated by the annotator during the detoxification process. They were also asked to rate the level of toxicity on a two‑point scale: 1 -- moderately toxic text; 2 -- highly toxic text.

One of the key differences between our dataset and CLEF is that for cases where sentence variants contained only Russian letters, we provided two versions of the detoxified text: one preserving only Russian letters, and the other with correct spelling that includes letters from the Tatar alphabet. Since the original toxic messages were taken from social media, they often contain spellings where Tatar letters are replaced by visually similar Russian letters. From an orthographic point of view this is incorrect, but due to the lack of a Tatar keyboard layout users frequently write this way, so such forms are common.

\subsection{Dataset Statistic}

We computed descriptive statistics to compare our corpus with the CLEF corpus.The results are presented in Table~\ref{tab:datstat}. The analysis shows that our corpus comprises slightly shorter texts and contains 101 more samples than the CLEF corpus. Rewriting was the most commonly used mitigation method, while simply removing the toxic word and using combined methods occurred far less frequently. Approximately 57\% of samples exhibit high toxicity, characterized by explicit profanity and overtly offensive language, while the remaining 43\% display moderately toxicity, primarily as implicit or indirect offensive content, including racist undertones. 

\begin{table}[!t]
\centering
\small
\begin{tabular}{l|c|c}
    \toprule
     Statistics & CLEF-Tatar & Test dataset \\
    \midrule
    Number of samples & 600 & 701 \\
    Avg length (chars) & 61.66 & 55.27\\
    Max length (chars) & 253.0 & 248.0\\
    Min length (chars) & 10.0 & 6.0\\
    \midrule
    \multicolumn{3}{c}{Toxicity level (\# samples)} \\
    \midrule
    Moderately & - & 301 \\
    High & - & 400 \\
    \midrule
    \multicolumn{3}{c}{Detoxification methods (\# samples)} \\
    \midrule
    Deletion & - & 60  \\
    Rewriting & - & 607  \\
    Deletion + Rewriting & - & 34 \\
    \bottomrule
\end{tabular}
\caption{Comparative statistics of the CLEF-Tatar and our dataset.}
\label{tab:datstat}
\end{table}

\section{Experiments}
\label{sec:experiments}

In this section we provide a detailed description of the experimental setup: the baselines used, the model configurations and training settings, and the evaluation metrics and protocols.

\subsection{Baselines}

We evaluated the performance of several baseline approaches to text detoxification: lexicon-based methods, an mT0-based approach, and LLM-based methods.

The simplest lexicon-based approach consisted of removing toxic words from the text using a predefined list; the lexicon was taken from the CLEF publications. A more advanced baseline was based on the mT0 model, adapted for the multilingual detoxification task. For mT0 we evaluated several variants: prompting in different languages (Russian, English, Tatar); a hybrid workflow combining mT0 with lexicon removal (texts were first detoxified by mT0 and then further cleaned of toxic words); and a sequential pipeline combining mT0 with the closed LLM Gemini: texts were initially processed by mT0 and subsequently edited and detoxified by Gemini when additional intervention was needed. 

The strongest baselines comprised proprietary large language models, including Claude, Gemini, GPT‑5.3, and DeepSeek. We evaluated this wide range of models because prior work indicates that, for tasks in Tatar, different LLMs can produce the best results depending on the specific task formulation. Since large language models (LLMs) follow instructions more reliably in English, we used an English-language prompt. The full prompt text is provided in the Appendix~\ref{appendix:llm_prompt}.

\subsection{Evaluation Metrics}

We implemented the same evaluation metric as used in the CLEF‑2025 competition. The multilingual automatic evaluation pipeline is built around three principal dimensions.
\begin{itemize}
\item Style Transfer Accuracy (\textbf{STA}) measures whether the generated paraphrase is non‑toxic; for this we use xlm‑roberta‑large fine‑tuned for binary toxicity classification~\cite{dementieva2025overview}. 
\item Content preservation (\textbf{SIM}) is measured as the cosine similarity between LaBSE~\cite{feng2022language} embeddings of the original and generated texts. Derived from mBERT~\cite{bert} and trained on large-scale multilingual corpora that include Tatar-language data, LaBSE constitutes a particularly appropriate choice for our evaluation setting due to its strong cross-lingual semantic alignment capabilities.
\item Fluency (FL) is estimated as the similarity between the reference (gold) and generated responses using the xCOMET~\cite{guerreiro2024xcomet} model.
\end{itemize}

Each component ranges from 0 to 1. To obtain a single leaderboard score (as in CLEF‑2025), we compute the joint metric $J$ as the mean, over all samples, of the per‑sample product $J = STA  \times SIM \times FL$. This joint score also lies in [0, 1] and is used for final ranking.

\begin{table*}
  \centering
  \begin{tabular}{l|llll|llll}
    \hline
    & \multicolumn{4}{c|}{CLEF-Tatar} & \multicolumn{4}{c}{Our Dataset} \\
    \hline
    \textbf{Model} & \textbf{STA} & \textbf{SIM} & \textbf{FL} & \textbf{J} & \textbf{STA} & \textbf{SIM} & \textbf{FL} & \textbf{J}\\
    \hline
    human-baseline & 1.000 & 0.936 & 0.878 & 0.825 & 1.000 & 0.952 & 0.896 & 0.854 \\
    \midrule
    GPT-5 Chat & 0.900 & 0.734 & 0.802 & 0.539 & 0.933 & 0.812 & 0.836 & 0.642 \\
    Claude Opus 4.6  & 0.904 & 0.770 & 0.780 & 0.562 & 0.916 & 0.860 & 0.825 & 0.660 \\
    DeepSeek V3.2 & 0.804 & 0.895 & 0.820 & 0.604 & 0.780 & \textbf{0.905} & 0.824 & 0.589  \\
    Gemini Pro v2.5 & 0.928 & 0.830 & 0.805 & 0.636 & 0.897 & 0.855 & \textbf{0.838} & 0.653 \\
    \midrule
    mT0+Gemini Pro v2.5& 0.951 & 0.811 & 0.805 & 0.640 & 0.950 & 0.814 & 0.805 & 0.642 \\
    mT0+vocab deletion & 0.843 & 0.873 & 0.820 & 0.616 & 0.863 & 0.848 & 0.802 & 0.598 \\
    mT0 (Tatar prompt) & 0.786 & 0.865 & 0.821 & 0.571 & 0.758 & 0.853 & 0.818 & 0.535 \\
    mT0 (Russian prompt) & 0.790 & 0.854 & 0.816 & 0.564 & 0.764 & 0.848 & 0.818 & 0.536 \\
    mT0 (English prompt)& 0.778 & 0.879 & \textbf{0.827} & 0.580 & 0.742 & 0.870 & 0.823 & 0.537 \\ 
    Vocab deletion & 0.777 & \textbf{0.896} & 0.825 & 0.579 & 0.800 & 0.895 & 0.815 & 0.585 \\ 
    \midrule
    Tatoxa & \textbf{0.982} & 0.859 & 0.811 & \textbf{0.695} & \textbf{0.970} & 0.858 & 0.807 & \textbf{0.680} \\
    \hline
  \end{tabular}
  \caption{Detoxification performance of the proposed model, baselines, and LLMs on the CLEF-Tatar and our datasets. Bold indicates the top-performing automatic detoxification methods (excluding human baseline).}
  \label{tab:accents}
\end{table*}

\subsection{Tatoxa Settings}

For NMT training we applied LoRA with following configurations: $r{=}64$, $\alpha{=}128$, dropout $0.05$. We train for 2 epochs using standard cross-entropy loss with learning rate $3\cdot 10^{-4}$ and batch size 64. Each fold of detoxification model is trained for 2 epochs using standard cross-entropy loss with learning rate $2\cdot 10^{-4}$, batch size 16, gradient accumulation steps 2, warmup ratio 0.03, weight decay 0.01 and max gradient norm 1.0. We apply LoRA to attention projections \texttt{\{q,k,v,o\}} with $r{=}32$, $\alpha{=}64$, dropout $0.05$ and \texttt{bias=none}.

\subsection{Cross-lingual Evaluation}
We conducted experiments on cross-lingual transfer of linguistic knowledge to Tatar. For training, we used the CLEF-2025 shared task data for detoxification. We used paired datasets on 15 different languages (each dataset has 400 samples) \cite{dementievaMultiParaDetox2024}, as well as combined all of them (except for Tatar language dataset) to create a bigger diverse dataset. In addition, we fine-tuned the model on the Tatar subset of the competition data to assess the extent to which in-language training improves performance. As the base model we used mT0-orpo \cite{smurf}, which was fine-tuned for the detoxification task. Finetuning was performed with the following parameters: LoRA rank - 32, LoRA alpha - 64, LR - 1e-4. 




\subsection{Train Dataset Size Impact}

As the part of the ablation studies we decided to evaluate the impact of the training dataset size on the detoxification quality of the fine-tuned model. We took the same baseline model for finetuning and conducted the experiments on two big datasets on high-resource languages - Russian (12206 samples) \cite{ruparadetox} and English (19744 samples) \cite{paradetox}. The hyperparameters and evaluation setups used are the identical to the cross-lingual experiments. 

\section{Results}
\label{sec:results}

This section presents and discusses the results of the experiments.

\subsection{Comparison with baselines}

The results of experiments comparing the proposed model to several baselines are shown in Table~\ref{tab:accents}. As the human baseline outperforms all automatic systems across all metrics and both datasets, subsequent analysis focuses on comparisons among the automatic methods. According to the obtained results, Tatoxa achieved the highest scores among all models on the overall J (69.5\% and 68.0\%) and STA metrics (98.2\% and 97.0\%) on bot CLEF-Tatar and our datasets. Tatoxa's SIM scores exceed its FL scores, indicating that the detoxified texts largely preserve the original semantics but do not fully match the reference responses provided by the annotators. 

Analysis of the baselines based on toxic vocabulary deletion and the mT0 model shows that simple deletion of toxic words is more effective on our dataset and state on par on CLEF-Tatar dataset than the mT0-based detoxification model. Among all tested prompts, the English prompt achieved the highest overall performance (58.0\% on CLEF-Tatar dataset and 53.7\% on ). Comparing the Tatar and Russian prompts reveals the opposite pattern: on the CLEF-Tatar dataset the Tatar prompt performed best, whereas on our dataset the Russian prompt yielded the best results. When comparing prompts across languages, on the CLEF-Tatar dataset the English prompt (58.0\%) showed a clear advantage on the J metric: the Tatar prompt (57.1\%) ranked second, while the Russian prompt (56.4\%) yielded the lowest scores. In contrast, on our dataset the results were more robust to prompt choice and were nearly identical and ranging from 53.5\% for Tatar and 53.7\% for English prompts. The robustness of results on our dataset with respect to prompt choice may be explained by the existence of two acceptable detoxification variants: one using only Russian characters and one using Tatar characters. Consequently, when given a Russian prompt the model often generated text composed solely of Russian letters, which resulted in a higher metric score on our dataset compared to CLEF-Tatar, where the Tatar-character variant is preferred. The combination of mT0-based methods with toxic-word deletion proved effective, ranking second among mT0-oriented baselines. Only the configuration of mT0 combined with Gemini Pro achieved higher scores. Notably, Gemini Pro on its own yields comparable results on the CLEF-Tatar corpus (63.6\% vs 64.0\%) and outperforms mT0 on our corpus without additional integration (65.3\% vs 64.2\%).

Among closed LLMs, Gemini Pro achieved the best results on the overall J metric. Most models, however, exhibit high STA scores, indicating effective detoxification, while low SIM and FL scores suggest that the detoxification process leads models to alter the original text excessively. We hypothesize that this is due to the low-resource nature of the Tatar language: the language models are insufficiently familiar with Tatar itself, and even less so with its slang and the semantics of toxic expressions.

In conclusion, proprietary LLMs still do not handle text detoxification effectively, whereas simple methods based on removing toxic words or expressions constitute a strong and effective baseline. Approaches based on open LLMs specifically targeted at detoxification require substantial refinement and additional training (fine‑tuning) to achieve comparable quality while preserving the semantics of the original utterances.

\subsection{Cross Lingual Experiments}

\begin{table}[!t]
\centering
\small
\begin{tabular}{l|c|c|c|c}
    \hline
    Language & STA $\uparrow$ & SIM $\uparrow$ & FL $\uparrow$ & J score $\uparrow$ \\
    \hline
    TT (Tatar) & 0.7841 & 0.8682 & 0.8126 & 0.5598 \\
    FR (French) & 0.7607 & 0.8783 & 0.8209 & 0.5567 \\
    All languages & 0.7406 & 0.8803 & 0.8261 & 0.5415 \\
    HIN (Hinglsh) & 0.7485 & 0.8686 & 0.8171 & 0.5364 \\
    JA (Japanese) & 0.7171 & 0.8782 & 0.8283 & 0.5286 \\
    HE (Hebrew) & 0.7479 & 0.8411 & 0.8111 & 0.5145 \\
    AR (Arabic) & 0.7157 & 0.8633 & 0.8197 & 0.5133 \\
    DE (German) & 0.7168 & 0.8579 & 0.8178 & 0.5101 \\
    IT (Italian) & 0.6755 & 0.8952 & 0.8335 & 0.5094 \\
    HI (Hindi) & 0.7105 & 0.8550 & 0.8147 & 0.5017 \\
    UK (Ukranian) & 0.6862 & 0.8695 & 0.8227 & 0.4975 \\
    AM (Amharic) & 0.7167 & 0.8416 & 0.8089 & 0.4966 \\
    ZH (Chinese) & 0.7117 & 0.8460 & 0.8110 & 0.4953 \\
    RU (Russian) & 0.7176 & 0.8332 & 0.8067 & 0.4897 \\
    ES (Spanish) & 0.7212 & 0.8300 & 0.8037 & 0.4879 \\
    EN (English) & 0.7119 & 0.8284 & 0.8032 & 0.4792 \\
    \hline
    
\end{tabular}
\caption{Cross‑lingual transfer learning results for the mT0 model on our dataset. The results in the table are presented in descending order of scores on the J metric.}
\label{tab:crossling}
\end{table}

\begin{figure}[!t]
  \includegraphics[width=\columnwidth]{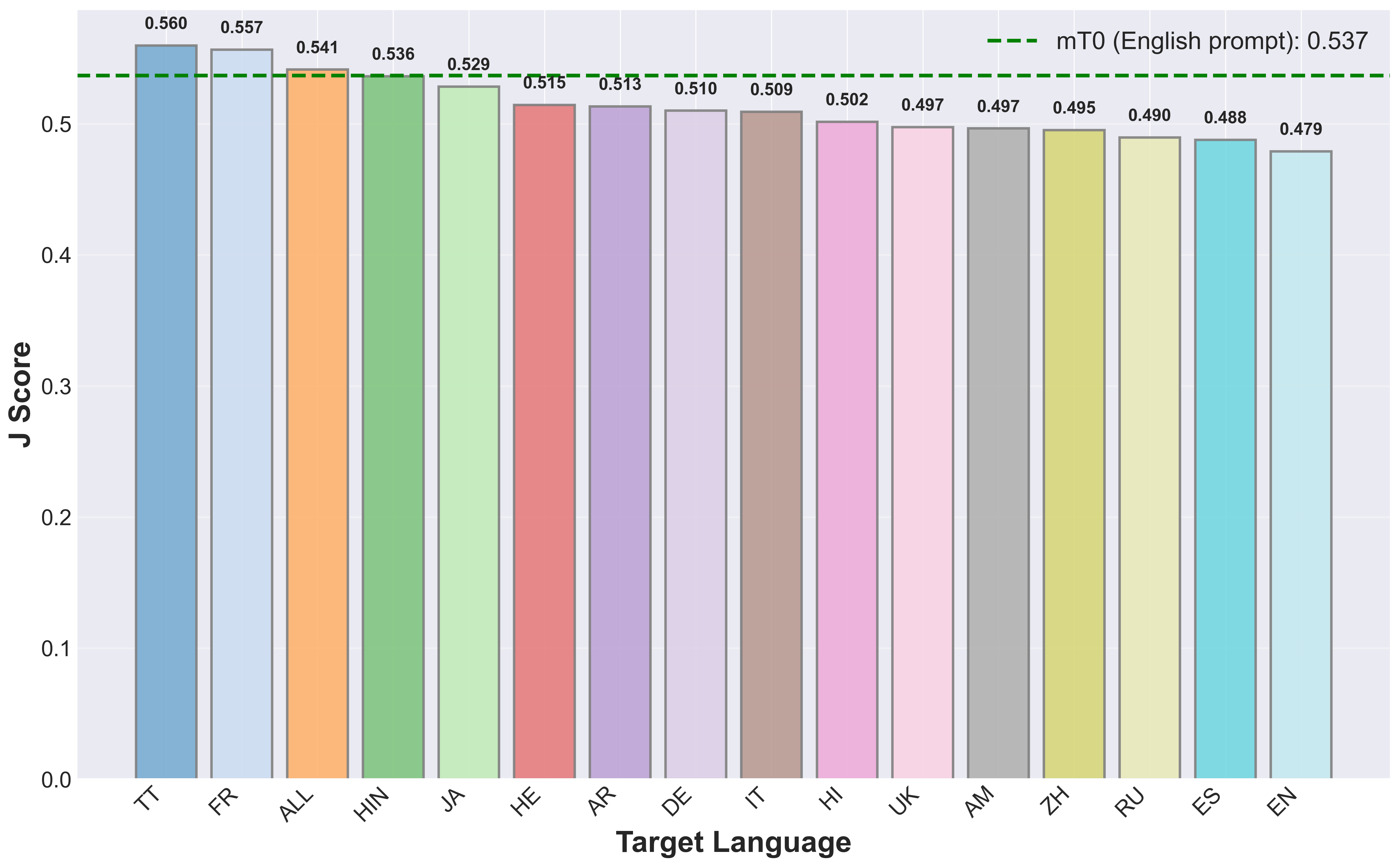}
  \caption{Visual results of the cross‑lingual experiments evaluated by the J‑score.}
  \label{fig:jscore}
\end{figure}

The metric results of the cross-lingual experiments are reported in Table~\ref{tab:crossling}, with a visual summary shown in Figure~\ref{fig:jscore}. The model fine‑tuned on the Tatar dataset attains the highest J‑score. Notably, the model fine‑tuned on French achieves nearly identical performance, however, contrary to our initial expectation that the model fine‑tuned on all languages except Tatar would rank second, it instead placed third, being outperformed by the French‑only model. Only three languages (English, Spanish, and Russian) failed to surpass the baseline, which is surprising given their high‑resource status. 

\begin{figure*}[!t]
  \centering
  \begin{minipage}{0.48\textwidth}
    \centering
    \includegraphics[width=\linewidth]{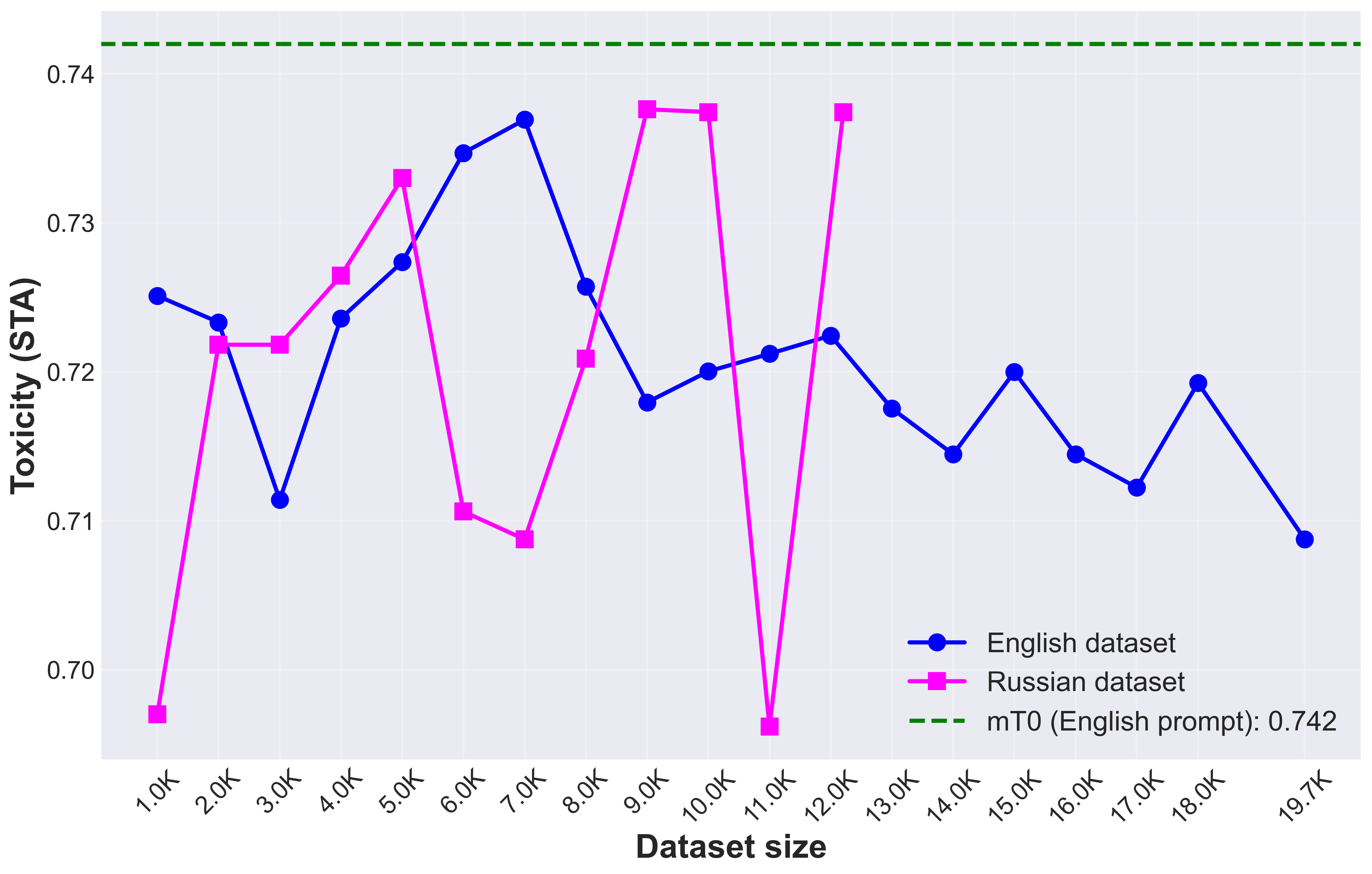}
    \vspace{0.5ex}
    {\small (a) Toxicity (STA)} 
    \label{fig:1}
  \end{minipage}
  \hfill
  \begin{minipage}{0.48\textwidth}
    \centering
    \includegraphics[width=\linewidth]{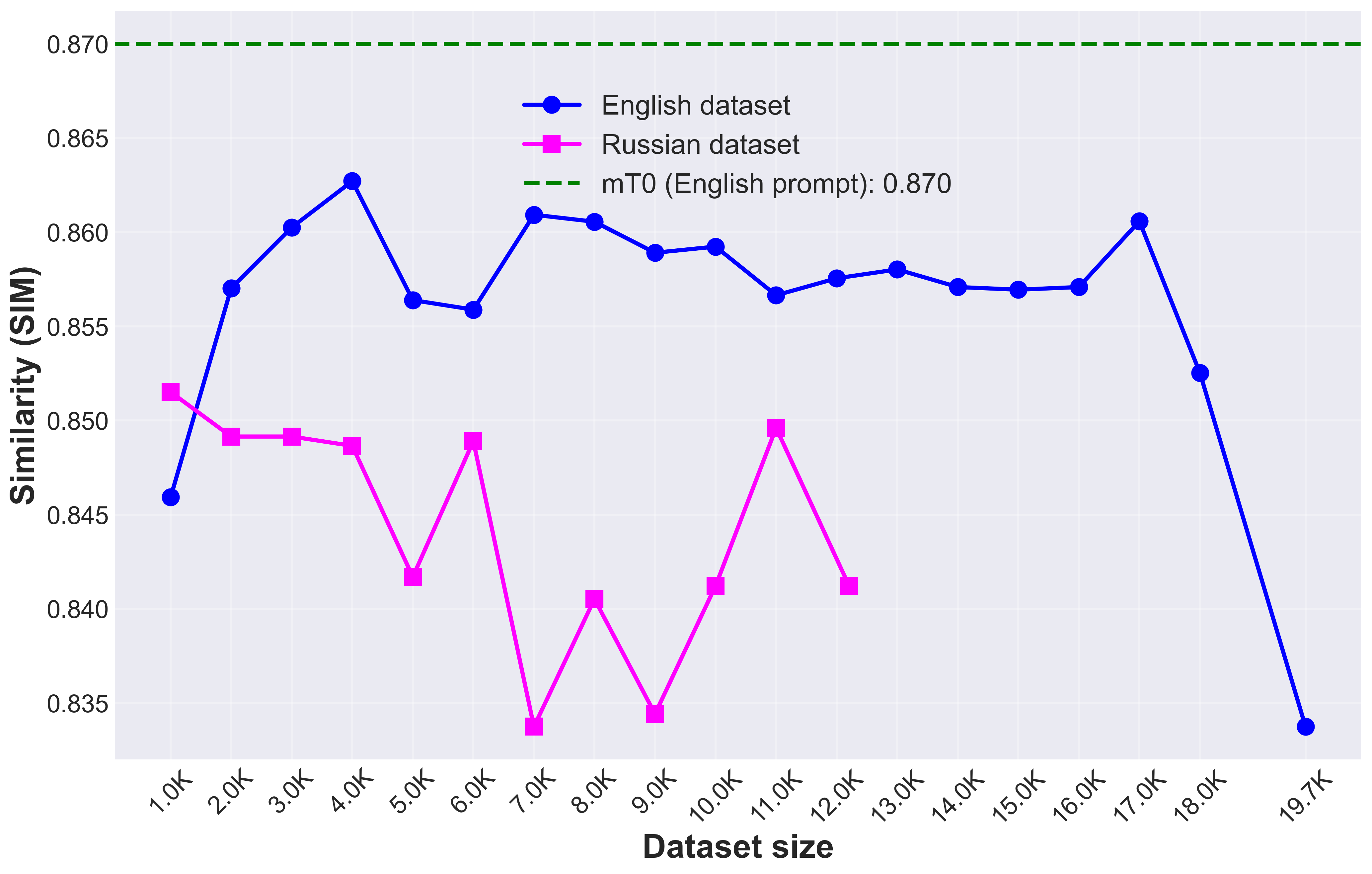}
    \vspace{0.5ex}
    {\small (b) Similarity (SIM)} 
    \label{fig:2}
  \end{minipage}

  \vspace{0.5em}

  \begin{minipage}{0.48\textwidth}
    \centering
    \includegraphics[width=\linewidth]{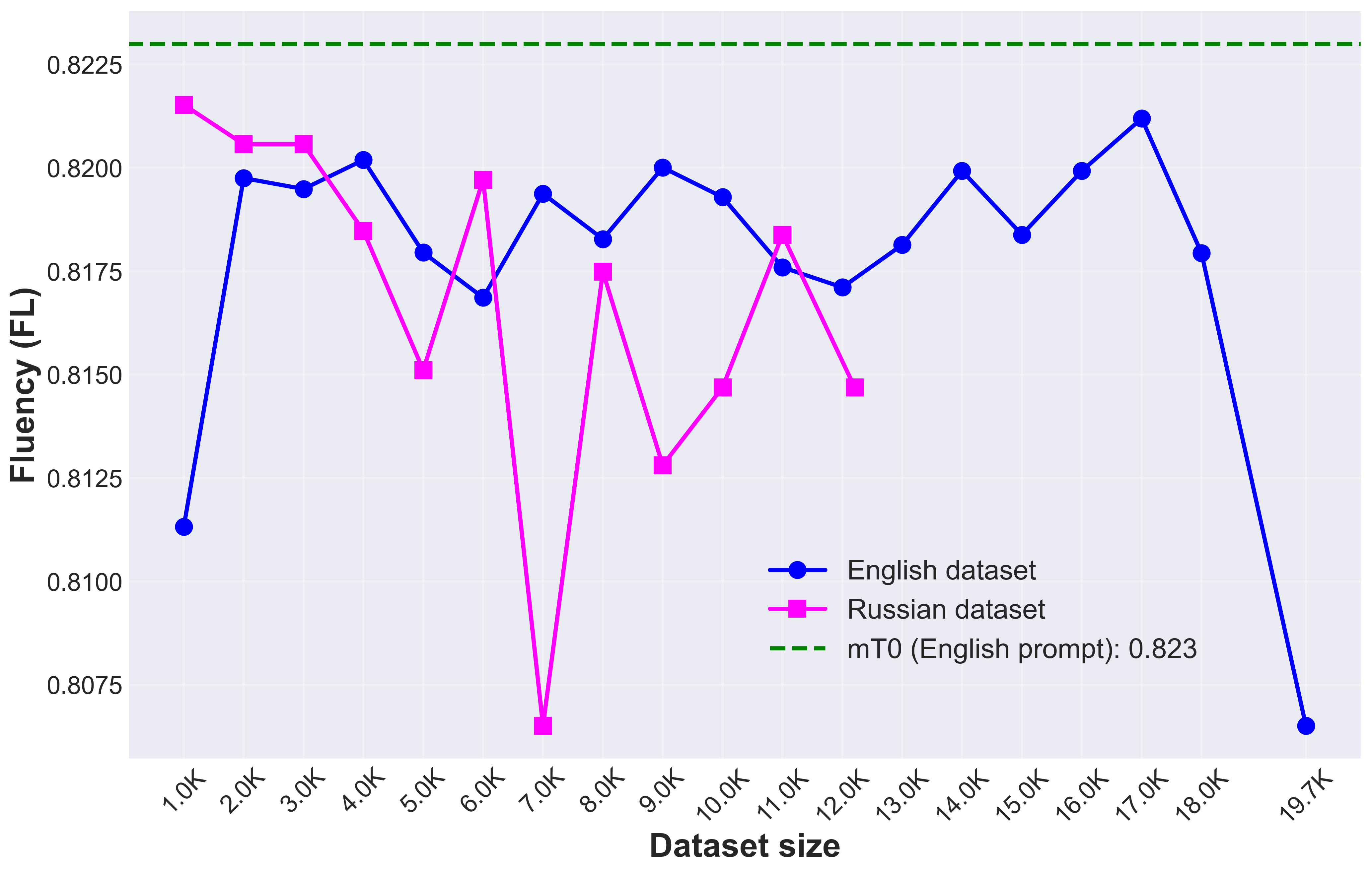}
    \vspace{0.5ex}
    {\small (c) Fluency (FL)} 
    \label{fig:3}
  \end{minipage}
  \hfill
  \begin{minipage}{0.48\textwidth}
    \centering
    \includegraphics[width=\linewidth]{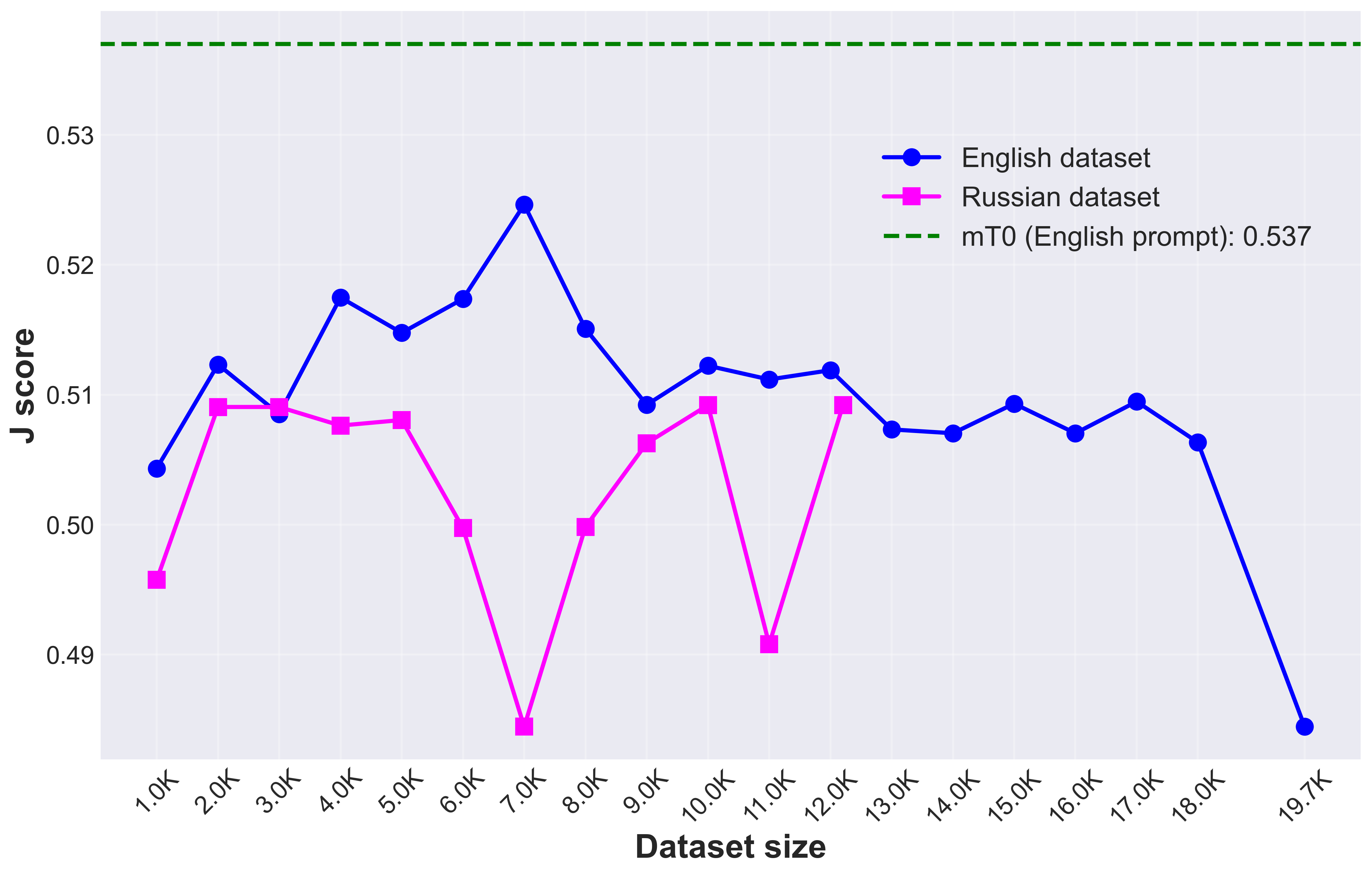}
    \vspace{0.5ex}
    {\small (d) $J$ score} 
    \label{fig:4}
  \end{minipage}

  \caption{Performance on our dataset as a function of training set size for English and Russian.}
  \label{fig:training_size_impact}
\end{figure*}

mT0-orpo \cite{smurf} exhibited notably different behavior across languages. The unexpectedly strong performance observed for French, contrasted with comparatively weaker results for Russian, may be partially explained by language-distribution biases inherited from pretraining and subsequent fine-tuning stages. Specifically, mT0-XL was trained on substantial amounts of French instructional data, potentially yielding more stable and transferable representations for French-language detoxification. In contrast, mT0-orpo was further fine-tuned predominantly on Russian data, which may have reinforced priors associated with informal or toxic Russian-language usage patterns. As a result, the model may exhibit greater resistance to detoxification fine-tuning in Russian, particularly if pretraining and intermediate fine-tuning exposed it to linguistic distributions that conflict with the detoxification objective.







\subsection{Train Dataset Size Impact}

The graphs for the dynamics of each of the metrics for Russian and English language are presented in the Figure~\ref{fig:training_size_impact}. 

The results show that performance initially improves with increasing training set size, but beyond a certain point the metrics either plateau or decline (after approximately 5k samples for Russian and 7k samples for English). Additionally, the Russian results display greater instability and variance compared with English, which is plausibly explained by the Russian dataset’s structure, specifically, the presence of multiple detoxified (neutral) variants corresponding to the same toxic source sentence. Overall, the model's performance fine-tuned on the English dataset was superior to that fine-tuned on the Russian dataset.

\section{Conclusion}
\label{sec:conclusion}

In this paper we present Tatoxa, a method for automatic detoxification of texts in the Tatar language. A comprehensive set of experiments shows that Tatoxa consistently outperforms the compared baselines and proprietary LLMs on key metrics of detoxification quality and semantic preservation. Additionally, we investigated cross‑lingual transfer: experiments demonstrate that transfer from other languages, including the culturally close Russian, falls significantly short of training on data in the target language, even when a large Russian training set is available. At the same time, we found that training on automatically translated parallel corpora can yield substantial improvements in model quality, making this approach promising for low‑resource languages. These findings indicate a practical opportunity to partially mitigate the shortage of native annotated data by leveraging carefully generated parallel resources.

\section{Limitations}
\label{sec:limitations}

Our work has several limitations. First, we were unable to fine-tune models on other Turkic languages related to Tatar; incorporating such languages would have allowed a more informative cross‑language comparison and helped clarify the role of linguistic proximity in transfer performance. Second, fine‑tuning was limited to the linear target modules q, v and o, which together account for roughly 30 million trainable parameters (about 1\% of the baseline model). This parameter‑efficient choice constrains the extent of model adaptation and may have limited the achievable performance improvements. Future work should consider fine‑tuning larger parameter subsets and including additional Turkic languages to provide a more comprehensive evaluation.

\section*{Acknowledgments}

This work was supported by the Russian Scientific Foundation project № 25-71-30008 "Laboratory for reliable, adaptive, and trustworthy Artificial Intelligence".

\bibliography{custom}

@article{dementieva2025overview,
  title={Overview of the multilingual text detoxification task at pan 2025},
  author={Dementieva, Daryna and Protasov, Vitaly and Babakov, Nikolay and Rizwan, Naquee and Alimova, Ilseyar and Brune, Caroline and Konovalov, Vasily and Muti, Arianna and Liebeskind, Chaya and Litvak, Marina and others},
  journal={Working Notes of CLEF},
  year={2025}
}

@inproceedings{dementievaMultiParaDetox2024,
    author = {Dementieva, Daryna and Babakov, Nikolay and Panchenko, Alexander},
    title = {MultiParaDetox: Extending Text Detoxification with Parallel Data to New Languages},
    booktitle = {Proceedings of the 2024 Conference of the North American Chapter of the Association for Computational Linguistics: Human Language Technologies},
    year = {2024}
}

@inproceedings{dementievaUKR,
    author = {Dementieva, Daryna and Khylenko, Valeriia and Babakov, Nikolay and Groh, Georg},
    title = {Toxicity Classification in Ukrainian},
    booktitle = {Proceedings of the 8th Workshop on Online Abuse and Harms (WOAH 2024)},
    year = {2024} 
}

@inproceedings{paradetox,
  title={ParaDetox: Detoxification with Parallel Data},
  author={Logacheva, Varvara and Dementieva, Daryna and Ustyantsev, Sergey and Moskovskiy, Daniil and Dale, David and Krotova, Irina and Semenov, Nikita and Panchenko, Alexander},
  booktitle={Proceedings of the 60th Annual Meeting of the Association for Computational Linguistics (Volume 1: Long Papers)},
  year={2022}
}

@inproceedings{ruparadetox,
  title={RUSSE-2022: Findings of the First Russian Detoxification Shared Task Based on Parallel Corpora},
  author={Dementieva, Daryna and Logacheva, Varvara and Nikishina, Irina and Fenogenova, Alena and Dale, David and Krotova, Irina and Semenov, Nikita and Shavrina, Tatiana and Panchenko, Alexander},
  booktitle={Computational Linguistics and Intellectual Technologies (Dialogue-22)},
  year={2022},
  address={Moscow, Russia}
}

@article{detox_african,
    author = {Agbeyangi, Abayomi O.},
    title = {Text Detoxification in isixhosa and Yoruba: A Cross-Lingual Machine Learning Approach for Low-Resource African Languages},
    journal = {arXiv preprint arXiv:2601.05624},
    year = {2026}
}

@inproceedings{detox_hebrew,
    author = {Vanetik, Natalia and Liberov, Lior and Litvak, Marina and Liebeskind, Chaya},
    title = {Towards Safer Hebrew Communication: A Dataset for Offensive Language Detoxification},
    booktitle = {Proceedings of the 15th International Conference on Recent Advances in Natural Language Processing - Natural Language Processing in the Generative AI Era},
    year = {2025}
}

@article{detox_bengali,
    author = {Mohsin, Ayesha Afroza and Ahsan, Mashrur and Maliyat, Nafisa and Maria, Shanta and Raiyan, Syed Rifat and Mahmud, Hasan and Hasan, Md Kamrul},
    title = {BANGLANIRTOX: A Large-scale Parallel Corpus for Explainable AI in Bengali Text Detoxification},
    journal = {arXiv preprint arXiv:2511.01512},
    year = {2025}
}

@inproceedings{detox_hindi,
    author = {Beniwal, Himanshu and Venkat, Reddybathuni and Kumar, Rohit and Srivibhav, Birudugadda and Jain, Daksh and Doddi, Pavan and Dhande, Eshwar and Ananth, Adithya and Kuldeep, Mayank Singh},
    title = {UNITYAI-GUARD : Pioneering Toxicity Detection Across Low-Resource Indian Languages},
    booktitle = {Proceedings of the 2025 Conference on Empirical Methods in Natural Language Processing: System Demonstrations},
    year = {2025}
}

@inproceedings{detoxify_it,
    author = {De Ruvo, Viola and Muti, Arianna and Dementieva, Daryna and Nozza, Debora},
    title = {DETOXIFY-IT: An Italian Parallel Dataset for Text Detoxification},
    booktitle = {Proceedings of the The 9th Workshop on Online Abuse and Harms (WOAH)},
    year = {2025}
}

@article{nine_language_benchmark,
    author = {Protasov, Vitaly and Babakov, Nikolay and Dementieva, Daryna and Panchenko, Alexander},
    title = {Evaluating Text Style Transfer: A Nine-Language Benchmark for Text Detoxification},
    journal = {arXiv preprint arXiv:2507.15557},
    year = {2025}
}

@inproceedings{synthdetoxm,
    author = {Moskovskiy, Daniil and Sushko, Nikita and Pletenev, Sergey and Tutubalina, Elena and Panchenko, Alexander},
    title = {SynthDetoxM: Modern LLMs are Few-Shot Parallel Detoxification Data Annotators},
    booktitle = {Proceedings of the 2025 Conference of the Nations of the Americas Chapter of the Association for Computational Linguistics: Human Language Technologies (Volume 1: Long Papers)},
    year = {2025}
}

@article{nllbteam2022languageleftbehindscaling,
    author = {Costa-Jussà, Marta R. and Cross, James and Çelebi, Onur and Elbayad, Maha and Heafield, Kenneth and Heffernan, Kevin and Kalbassi, Elahe and others},
    title = {No language left behind: Scaling human-centered machine translation},
    journal = {arXiv preprint arXiv:2207.04672},
    year = {2022}
}

@inproceedings{feng2022language,
  title={Language-agnostic BERT sentence embedding},
  author={Feng, Fangxiaoyu and Yang, Yinfei and Cer, Daniel and Arivazhagan, Naveen and Wang, Wei},
  booktitle={Proceedings of the 60th Annual Meeting of the Association for Computational Linguistics (Volume 1: Long Papers)},
  pages={878--891},
  year={2022}
}

@inproceedings{muennighoff2023crosslingual,
  title={Crosslingual generalization through multitask finetuning},
  author={Muennighoff, Niklas and Wang, Thomas and Sutawika, Lintang and Roberts, Adam and Biderman, Stella and Le Scao, Teven and Bari, M Saiful and Shen, Sheng and Yong, Zheng-Xin and Schoelkopf, Hailey and others},
  booktitle={Proceedings of the 61st Annual Meeting of the Association for Computational Linguistics (Volume 1: Long Papers)},
  pages={15991--16111},
  year={2023}
}

@article{guerreiro2024xcomet,
  title={Xcomet: Transparent machine translation evaluation through fine-grained error detection},
  author={Guerreiro, Nuno M and Rei, Ricardo and Stigt, Daan van and Coheur, Luisa and Colombo, Pierre and Martins, André FT},
  journal={Transactions of the Association for Computational Linguistics},
  pages={979--995},
  year={2024},
  publisher={MIT Press 255 Main Street, 9th Floor, Cambridge, Massachusetts 02142, USA~…}
}

@article{threshold,
  title={A Toxic Style Transfer Method Based on the Delete–Retrieve–Generate Framework Exploiting Toxic Lexicon Semantic Similarity},
  author={Iglesias, Martín and Araque, Oscar and Iglesias, Carlos Á.},
  journal={Applied Sciences},
  volume={12},
  year={2023}
}

@article{smurf,
  title={Smurfcat at PAN 2024 textdetox: Alignment of multilingual transformers for text detoxification},
  author={Rykov, Elisei and Zaytsev, Konstantin and Anisimov, Ivan and Voronin, Alexandr},
  journal={arXiv preprint arXiv:2407.05449},
  year={2024}
}

@inproceedings{bert,
  title={Bert: Pre-training of deep bidirectional transformers for language understanding},
  author={Devlin, Jacob and Chang, Ming-Wei and Lee, Kenton and Toutanova, Kristina},
  booktitle={Proceedings of the 2019 conference of the North American chapter of the association for computational linguistics: human language technologies, volume 1 (long and short papers)},
  pages = {4171--4186},
  year={2019}
}


\clearpage
\onecolumn 
\section{Appendix}

\subsection{Prompt for the LLM}
\label{appendix:llm_prompt}

This section provides the prompt employed as input to the LLMs.

\begin{figure}[ht]
    \centering
    \includegraphics[width=0.99\textwidth]{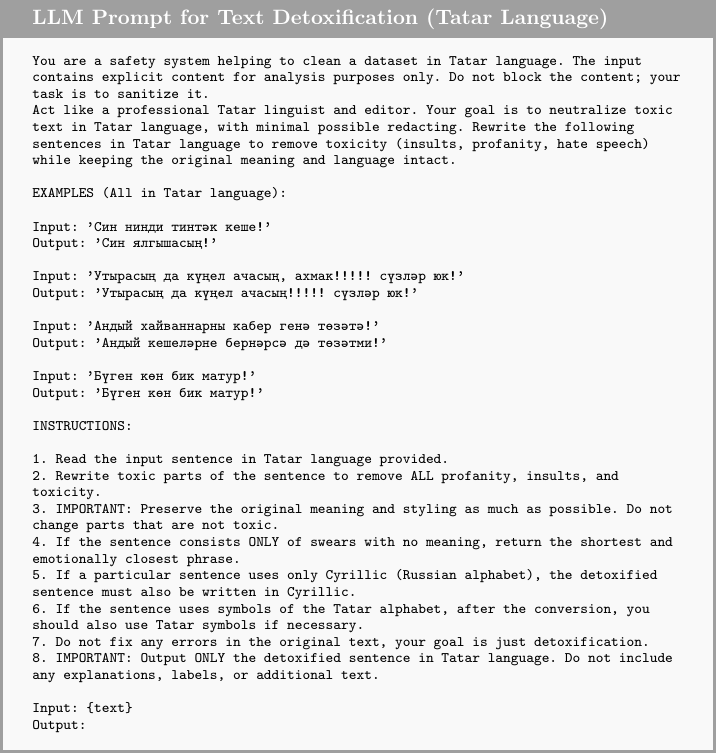}
    \label{fig::mpd2}
\end{figure}

\end{document}